# Closed-loop control of seizure activity via real-time seizure forecasting by reservoir neuromorphic computing

**Maryam Sadeghi[1†], Darío Fernández-Khatiboun[1†], Yasser Rezaeiyan[1,4], Saima Rizwan[2], Alessandro Barcellona[3], Andrea Merello[3], Marco Crepaldi[3], Gabriella Panuccio[2*], Farshad Moradi[1,4*]**

[1] Department of Electrical and Computer Engineering, Aarhus University, Aarhus, 8200, Denmark
[2] Enhanced Regenerative Medicine Lab, Istituto Italiano di Tecnologia, Genova, 16163, Italy
[3] Electronic Design Laboratory, Istituto Italiano di Tecnologia, Genova, 16152, Italy
[4] ICELAB, SDU Microelectronics, University of Southern Denmark, Denmark

[†]These authors contributed equally to this work.
*Author to whom any correspondence should be addressed.

E-mail: gabriella.panuccio@iit.it
moradi@ece.au.dk

## Abstract

Closed-loop brain stimulation holds potential as personalized treatment for drug-resistant epilepsy (DRE) but still suffers from limitations that result in highly variable efficacy. First, stimulation is typically delivered upon detection of the seizure to abort rather than prevent it; second, the stimulation parameters are established by trial and error, requiring lengthy rounds of fine-tuning, which delay steady-state therapeutic efficacy. Here, we address these limitations by leveraging the potential of neuromorphic computing. We present a neuromorphic reservoir computing hardware system capable of driving real-time personalized free-run stimulations based on seizure forecasting, wherein each forecast triggers an electrical pulse rather than an arbitrarily predefined fixed-frequency stimulus train. The system achieves 83.33% accuracy in forecasting seizure occurrences during the training phase. We validate the system using hippocampal spheroids coupled to 3D microelectrode array as a simplified testbed, achieving seizure reduction >97% during the real-time processing while primarily using instantaneous stimulation frequencies within 20 Hz, well below what typically used in clinical practice. Our work demonstrates the potential of neuromorphic systems as a next-generation neuromodulation strategy for personalized DRE treatment, leveraging their sparse and event-driven processing for real-time applications.

## Introduction

Epilepsy is a chronic neurological disorder characterized by the persistent predisposition of the brain to generate seizures, i.e., sudden excessive electrical discharges due to hyperexcitability (1). Epilepsy affects 70 million people globally (2, 3) and carries among the highest global burden within neurological disorders (1). Between 30 to 40% of patients do not respond to anti-seizure medications (2) and are therefore diagnosed with drug-resistant epilepsy (DRE). For the latter, the current first-line treatment is the surgical resection of the epileptic focus; however, this is a no-return option that is only feasible in a subset of patients and cannot guarantee complete freedom from seizures.

Brain stimulation emerges as a valid alternative to treat DRE, as it is safe and reversible (4, 5). However, the stimulation protocols still rely on arbitrary choices made by trial and error among a nearly infinite combination of stimulation parameters (6, 7). As a result, currently achievable therapeutic outcomes are promising (8, 9), but there is still room for improvement.

Closed-loop systems, which operate based on feedback from ongoing brain activity, are at the forefront of personalized brain stimulation for DRE treatment as they are designed to deliver electrotherapy when it's most needed. The Responsive Neuro Stimulation (RNS) system is the emblem of such closed-loop devices, and it is currently the only one approved by the Food and Drug Administration. The RNS (and similarly other closed-loop devices still in preclinical research) is primarily designed to respond to the detection of a seizure so to abort it at its onset (10). However, this approach is sub-optimal for two interrelated reasons: (i) ideally, the seizure should be prevented rather than aborted and (ii) once the seizure has started, it might be difficult to entrain the epileptogenic network. Particularly, cases of sudden unexpected death in epilepsy can still happen (11) although at a reduced incidence (3).

Seizure prediction has long been considered the method of choice to provide preventive stimulation therapy. However, it is a deterministic approach in that it aims at predicting the seizure onset within an arbitrarily defined time window (i.e., the prediction horizon). As such, it disregards that seizure generation is a dynamical process that does not follow a linear time course; thus, stimulation within a fixed time window before the seizure may yield opposite effects depending on the dynamical state of the epileptogenic network at the time of stimulation (12). Seizure forecasting, a probabilistic approach that continuously assesses the likelihood of seizure occurrence (13), is emerging as a better approach to inform the optimal stimulation timing throughout the interictal-to-ictal continuum. Still, the lack of well-defined electrographic biomarkers to obtain such likelihood estimation in real time prevents leveraging the full potential of seizure forecasting for personalized preventive stimulation. Machine learning-based approaches promise to overcome this limitation (13, 14, 15), but they come at the cost of power- and computation-hungry devices. In this framework, neuromorphic ('brain-inspired') systems are gaining momentum for their potential in low-power computation-efficient biomedical applications, including signal processing and neuromodulation, thanks to their appealing potential to replicate the energy-efficient functionality of the human brain and emulate the information processing of biological neurons (16, 17). Particularly, neuromorphic systems permit encoding information using spike trains (18), thus enabling event-driven computation. Moreover, the spatiotemporal sparsity of their neural activity supports implementing highly energy-efficient systems (19). Building on this concept, we have designed and implemented a neuromorphic computing system (NCS) on FPGA leveraging reservoir computing and trained it for real-time seizure forecasting supporting closed-loop electrical stimulation to prevent seizures.

Figure 1 illustrates the closed-loop architecture for electrical stimulation driven by seizure forecasting operated by the NCS (see Methods for full details).

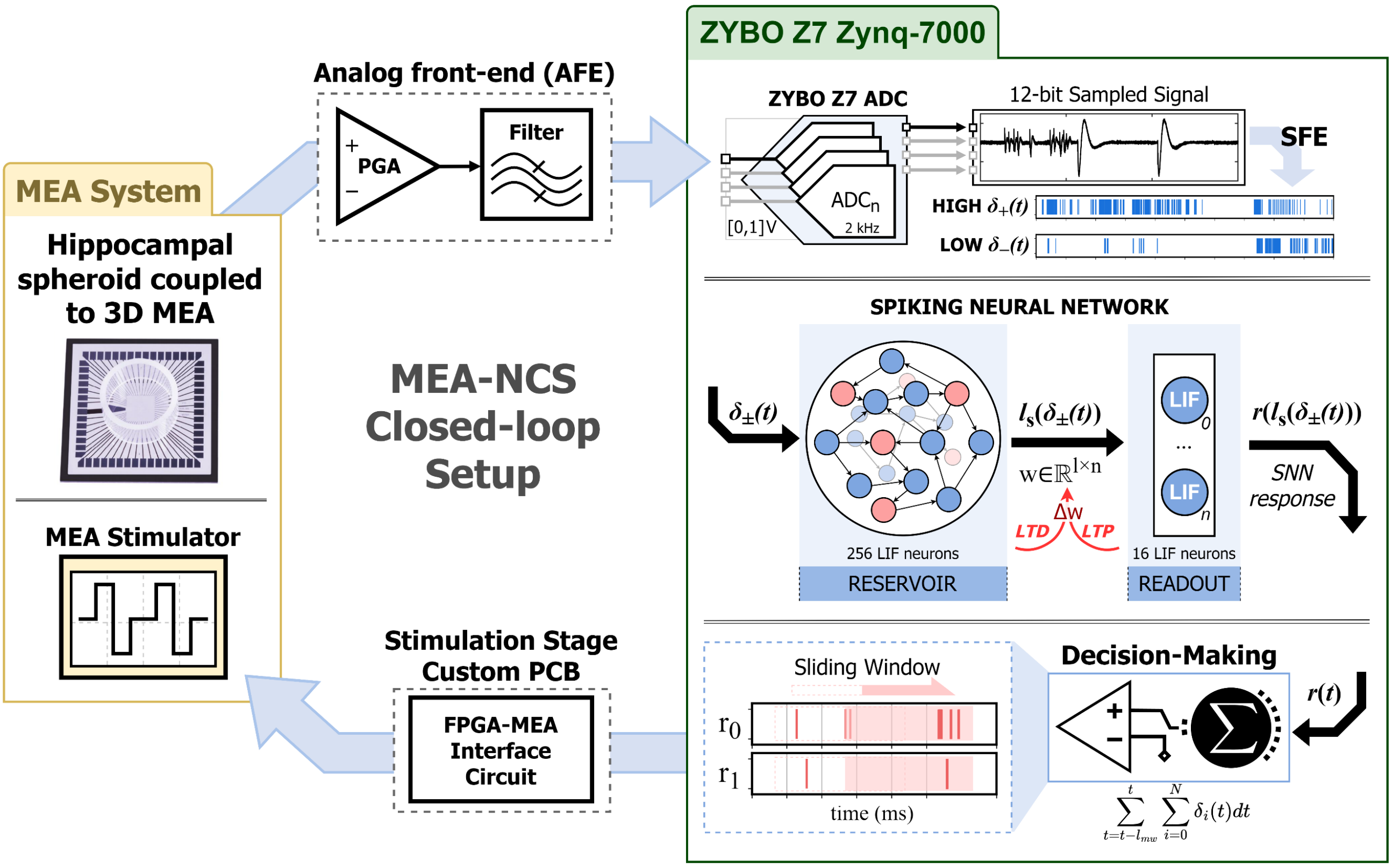

**Figure 1. Closed-loop operation of the neuromorphic computing system for seizure control in hippocampal spheroids.** Signals recorded from a rodent hippocampal spheroid using a 3D MEA are amplified by the analog front-end (AFE), producing a voltage signal in the range of 0 to 1 V. The signal is then digitized by the FPGA's ADC and converted into spike streams by a spike encoder. The spike data is processed through the SNN, composed of reservoir and readout layers. The results from the SNN are processed by the decision-making block to forecasts seizures. Positive forecasts then trigger stimulation to the MEA stimulator to prevent seizure activity.

The architecture comprises commercial microelectrode array (MEA) electrophysiology equipment, interfaced with the FPGA-implemented NCS via a custom interface PCB. As a simplified biological testbed, we used rodent primary hippocampal spheroids, which spontaneously generate epileptiform activity consisting of ictal (seizure) and interictal (between seizures) events (20). Signals are captured from a spheroid coupled to a 3D MEA and subsequently amplified in real-time using an adjustable multi-channel analog front-end to enhance signal quality for further processing. The amplified signals are then digitized via 12-bit FPGA's ADCs at a sampling frequency $f_{in}$ of 2 kHz and streamed into a spike threshold encoder, which converts them into spike streams based on the input dynamics. These streams are processed sequentially, in real time, by the NCS, where the system processes the input before the next sample is received. The NCS follows a liquid state machine (LSM) architecture consisting of two layers of spiking neurons. The first layer, a reservoir layer, uses randomly assigned recurrent synaptic connections and weights, creating recurrent pathways to perform complex nonlinear transformations and map the input data into a high-dimensional output space while extracting temporal features in real-time. The second layer is a trainable readout layer consisting of 2 neurons, which receive the output of the reservoir. This readout layer employs a reward-modulated spike-timing-dependent plasticity (R-STDP) algorithm (21), a supervised learning approach that enables real-time training. Finally, a decision-making block interprets the readout layer's firing activity to determine whether to activate the built-in stimulator of the MEA system (see Methods). This yields an ad-hoc stimulation pattern mirroring the timings of the true states in the decision-making block, i.e., each true state triggers a single electrical pulse instead of activating a stimulus train of predefined frequency and duration.

Using rodent hippocampal spheroids as a simplified in vitro testbed, we collected neural activity data to evaluate the NCS. The neural network from the NCS is first trained offline using approximately 70 segments of recorded data, achieving an accuracy of 83.33% in forecasting seizure states during the train phase. After training, the NCS is initialized with the synaptic weights obtained during the training process and coupled to the MEA electrophysiology system to process streams of data from rodent hippocampal spheroids and trigger stimulation upon detection of epileptiform occurrences that may precede an ictal event. The NCS demonstrates a reduction of seizure activity by more than 97% while delivering ad-hoc patterned stimulation, operating predominantly in the low-frequency stimulation range. Our work paves the way for a new generation of power- and computation-efficient neuromodulation devices grounded on seizure forecasting to treat DRE.

# Materials and Methods

## Experimental Design

Our goal was to obtain an NCS capable of real-time seizure forecasting that would drive electrical stimulation through its forecasts to prevent seizure activity. To this end, our experimental design comprises training, testing and validation sessions.

The NCS training dataset consisted of labelled pre-recorded signals from n = 6 hippocampal spheroids including interictal and ictal discharges, collected for the initial characterization of the epileptiform patterns generated by hippocampal spheroids (20). The pre-trained NCS was then tested online using pre-recorded signals injected in saline as voltage stimulus waveform. This step served two purposes: (i) validating the correct communication of the NCS with the electrophysiology set-up and (ii) obtaining an initial list of values of the NCS parameters influencing its behavior. Lastly, we validated the NCS in closed-loop stimulation experiments using hippocampal spheroids, according to the experimental protocol depicted in Fig. 3A. The effect of the NCS-driven stimulation in terms of ictal activity reduction was quantified with respect to the reference quantification of ictal activity during pre-stimulus baseline recording. The post-stimulus recording served as a further control to address possible after-effects that might indicate neuroplasticity.

## Hippocampal Spheroids Preparation and Maintenance

Hippocampal spheroids were prepared from Sprague-Dawley rat embryonic hippocampi harvested at embryonic age E17.5 and processed as detailed in (20). For each spheroid, 30,000 cells were seeded in each well of an ultra-low adhesive U-shaped 96-well plate (Cell Carrier™ -96, PerkinElmer) with 200 ml of Neurobasal medium supplemented with 2% B27, 1% PenStrep and 1% Glutamax. On the third day after seeding, the spheroids were transferred to a 24-well plate coated with 3% agarose (UV-treated for 15-20 min), one spheroid per well. Neurobasal medium was replaced weekly by 50%. Spheroids were maintained in a cell culture incubator at 37°C and 5% $CO_2$. Reagents were purchased from Thermo Fisher, Italy.

## Microelectrode Array Electrophysiology

Extracellular field potential recordings were performed acutely via 3D MEA (8x8 layout, TiN electrodes, diameter 12 µm, height 80 µm, inter-electrode distance 200 µm, impedance ~150 kΩ, internal reference electrode) using the MEA2100-mini-HS60 amplifier connected to the IFB v3.0 multiboot interface board through the SCU signal collector unit. Signals were sampled at 5 kHz (lowpass filtered at 2 kHz before digitization), acquired via the Multichannel Experimenter software and stored in the PC hard drive for offline analysis. All recordings were performed in artificial cerebrospinal fluid (ACSF) composed of (mM): NaCl 117, KCl 3.75, $KH_2PO_4$, 1.25, $MgSO_4$ 0.5, $CaCl_2$ 2.5, D-glucose 25, $NaHCO_3$ 26, L-Ascorbic Acid 1. The ACSF was maintained at ~37° C with the use of a custom-made heating lid covering the headstage, along with the warming of the MEA amplifier base, and equilibrated at pH ~7.4 through humidified carbogen delivered via a tubing connected to the heating lid. The recording temperature was controlled by a TC02 thermostat and checked with a k-type thermocouple in all experiments.

Each spheroid was taken from the cell culture incubator immediately before the experimental session and placed onto the 3D MEA, where it was let habituate for 20-30 minutes until steady-state activity before recording. The equipment for MEA electrophysiology and temperature control was purchased from Multichannel Systems (MCS), Reutlingen, Germany. Chemicals for the ACSF were purchased from Sigma-Aldrich, Italy.

## Dataset preparation and training process

To test and optimize the NCS before physical deployment, we simulated the SNN behavior using a co-simulation environment with the NEST simulator framework and Spiking Jelly 0.0.14 in Python 3.11.9, allowing precise replication of the hardware functionality. The simulation environment is quantified using the same bit precision as the parameters implemented on the FPGA hardware, enabling a 1:1 correspondence between the simulated and trained model, and the deployed system.

As described in (20), the epileptiform phenotype of these spheroids may be purely interictal (i.e., no ictal activity) or comprise both interictal and ictal events. This behavior suggests that the information conveyed by interictal events carries different values for seizure forecasting. Thus, distinguishing these events may support the design of robust seizure forecasting systems. To this end, we trained a first neuron to become sensitive to interictal discharges coexisting with ictal activity, and a second neuron to respond to interictal discharges appearing in isolation (i.e., in spheroids not generating ictal activity). The dataset is a subset of what is described in (20) and its preparation is described in Figure 2. It consists of ten 20-minute MEA recordings from different hippocampal spheroids, originally digitized at 20 kHz and resampled for this work for both the FPGA ADC and the NCS speed limitations at $f_{in}$ = 2 kHz. The dataset organizes the recorded discharges into labelled 20-second segments based on the recordings. Ictal discharges are then removed from the recordings, dissipating their temporal relation with the interictal events and, at the same time, the interictal time series itself to enhance the robustness of the training process. In summary, the dataset combines both baseline and interictal activity from both types of electrical phenotypes generated by the spheroids, with each interictal segment labelled accordingly (i.e., ‘ictal’ and ‘non-ictal’). Additionally, following the application of a modified SFE encoding algorithm (detailed in the next section of this work), we encode the information into spiking segments. Those segments exhibiting a low firing rate of <1% firing rate are excluded from the dataset to eliminate samples that would have negligible impact on the system's activity, speeding up the training process.

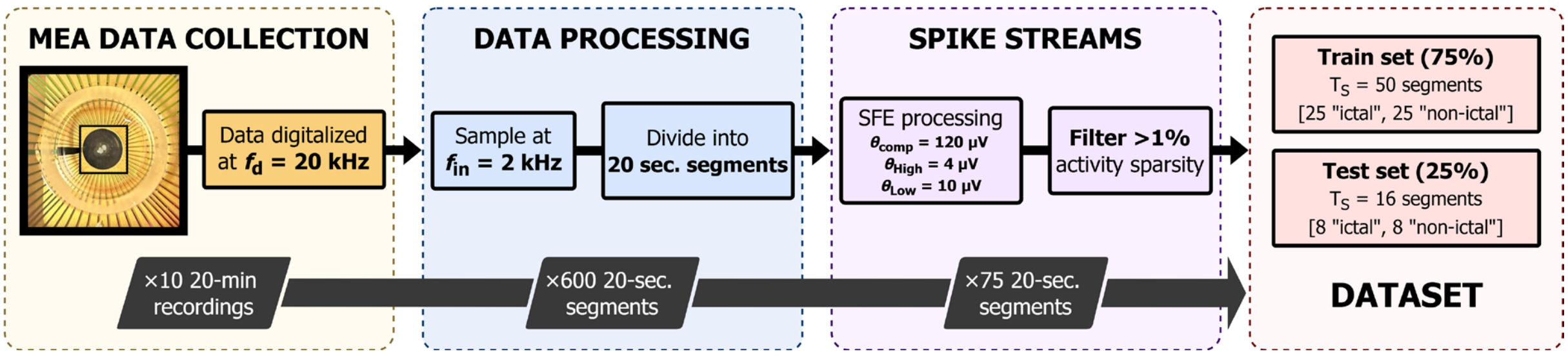

**Figure 2. Dataset pipeline preparation.** The data preparation starts with 20-minute recording data collected from the MEA system and processed through the same pipeline implemented in the FPGA to simulate the same scenario implemented in the FPGA for processing spheroids in real-time. Additionally, the data sampled at $f_{in}$ = 2 kHz is divided into 20-seconds segments to evaluate and optimize the training process. Furthermore, the spike segments are filtered based on their firing activity to reduce both train and test datasets with only the relevant spiking activity.

The complete training set is composed of 50 random samples of the spheroid's activity, evenly divided between "ictal" and "non-ictal" classes. The NCS is trained offline, where the NCS processes each sample updating the synaptic weights dynamically in a local fashion based on spike-timing relations between postsynaptic and presynaptic spikes of the R-STDP learning algorithm (detailed in another section). The training phase consisted of 50 epochs, with weight adjustments occurring at each timestep to reinforce neuronal activity. To evaluate the model's performance during the training process, we compute the accuracy of the system using samples from the training process every 5 epochs. At the end of the training process, we use a separate testing set composed of 16 20-second segments to ensure no overlapping with training data. Once the training process is validated, the architecture is deployed in inference-mode only, where their parameters, architecture, and the synaptic weights are uploaded into the FPGA to operate the system in real time with data streamed from the MEA system.

## Epileptiform signals capturing and analog front-end custom circuitry

Figure 3 shows a simplified schematic of the analog front-end (AFE) designed, which consists of a differential readout channel, a blanking circuit, a closed loop driven right leg (DRL) circuit, and a digital offset cancellation loop.

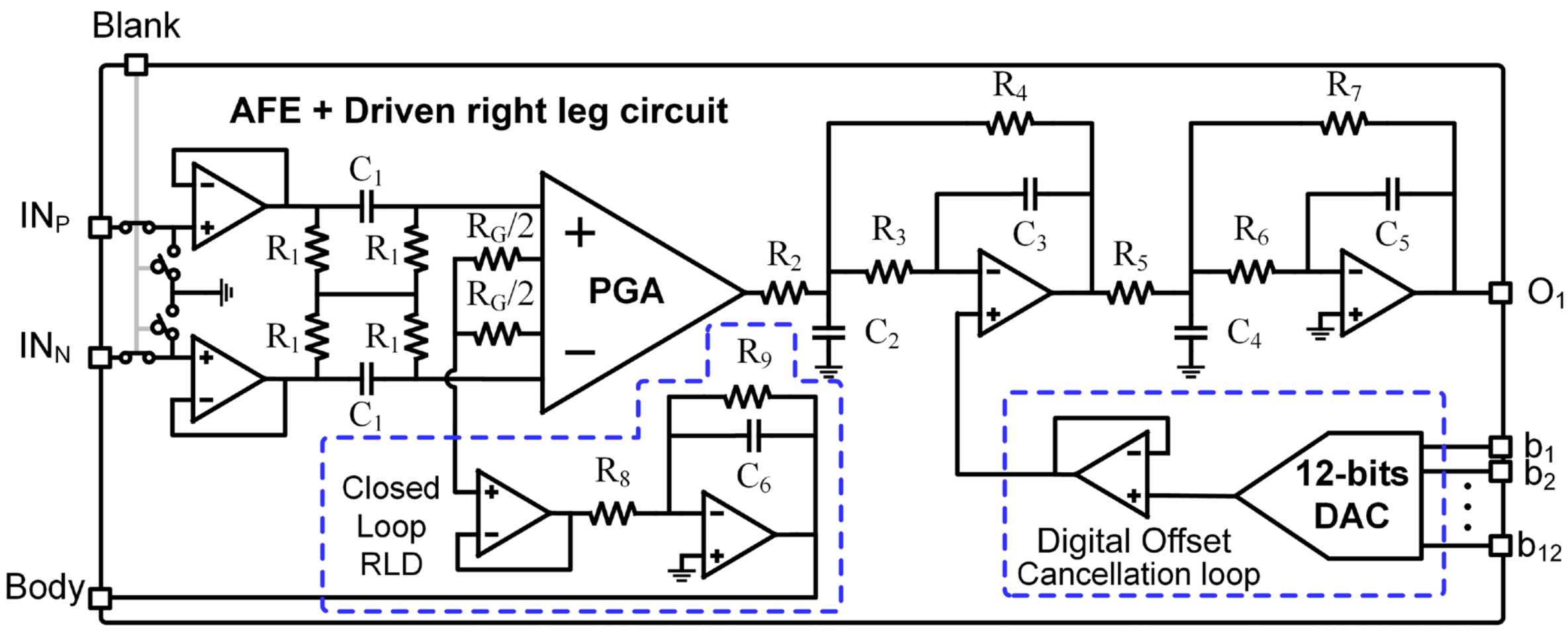

**Figure 3. Front-end architecture and spike encoding scheme.** The AFE processes the data stream from the MEA systems and amplifies to a range of 0-1V so it can be processed by the FPGA ADCs.

At the input stage, two buffers increase the input impedance of the front-end and enhance its common-mode rejection ratio (CMRR). The buffered signals are fed into an AC-coupled instrumentation amplifier (IA) to remove the electrodes' offset and avoid the amplifier's saturation. Furthermore, a ground-isolated balanced AC-coupling circuit improves the CMRR and eliminates direct signal paths from the IA input to the grounds. This circuit ensures AC coupling for the differential signal while providing a biasing path for the IA. The gain of the IA is adjustable between 20dB-60dB using the $R_G$ resistor. Following the IA, a fourth-order Butterworth low-pass filter is employed, offering a bandwidth of 10 kHz which enables the sensing of both action potential and local field potential. This filter achieves >75 dB attenuation in the stopband at 100 kHz and maintains passband ripple below 1 dB, ensuring high signal fidelity.

The AFE is biased such that the output common-mode voltage is set to half the ADC's full-scale range, optimizing the output swing. However, the IA's offset variations can shift the output common-mode voltage at high gain settings. To mitigate this, the digital offset cancellation loop stabilizes the output common-mode voltage at 0.5 V, corresponding to the midpoint of the ADC's range. At the start of channel activation, the input channel is temporarily shorted to ground, and the output voltage is monitored by the FPGA's ADC. If the common-mode voltage deviates from 0.5 V, the DAC output is adjusted to realign the output common-mode voltage to the desired 0.5 V level. This ensures reliable operation and prevents offset-induced distortions.

## Modified SFE encoding algorithm

SFE encoding algorithm (24) computes the variation of a continuous signal $c(t)$ to generate positive and negative spikes when the variation of amplitude surpasses a threshold $\theta_{SFE}$. This threshold is fixed, which may not capture specific patterns at different ranges of amplitude that a signal may have. To solve such an issue, we proposed a modified version of the SFE algorithm. In this version, depending on the current value of the signal with respect to the threshold comparator $\theta_{comp}$, the SFE generates spike events on two separate channels, HIGH and LOW, each with a different and fixed threshold $\theta_{high}$ or $\theta_{low}$. This way, we can differentiate between frequency components based on the amplitude of the signal. Both the pseudocode for the modified SFE used in the proposed NCS deployed in the FPGA and the conventional SFE are shown in Figure 4. Observe how the proposed SFE includes two output channels and a set of parameters $\theta_{\{comp,\ high,\ low\}}$. The configuration of the SFE algorithm and its parameters is achieved by analyzing the combined performance of the NCS and the SFE. We perform a series of ISI analysis and spike averaging count per sample. By limiting the activation and firing rate of the system we ensure stable behavior within the recurrent dynamics of the reservoir and reduced power consumption due to the reduced samples to be processed within each signal. The configuration used in the NCS is displayed in TABLE I, while the studies are collected in the Results section of this work.

**Algorithm 1** Modified Step Forward Encoding (SFE) Algorithm

**Input:** Continuous signal: $c(t)$, Comparator threshold: $\theta_{comp}$, HIGH threshold: $\theta_{high}$, LOW threshold: $\theta_{low}$
**Output:** HIGH Channel: $s_{high}(t)$, LOW Channel: $s_{low}(t)$

```
Initialize s_high(t) ← zeros of length N = length of s(t)
Initialize s_low(t) ← zeros of length N
s(t) ← c(0)
b_high ← s(0)
b_low ← s(0)
for each timestep i do
    s(t) ← c(t)
    if |s(t)| > θ_comp then
        if s(t) > b_high + θ_high then
            s_high(t) ← 1
            b_high ← b_high + θ_high
        else if s(t) < b_high − θ_high then
            s_high(t) ← −1
            b_high ← b_high − θ_high
        end if
    else
        if s(t) > b_low + θ_low then
            s_low(t) ← 1
            b_low ← b_low + θ_low
        else if s(t) < b_low − θ_low then
            s_low(t) ← −1
            b_low ← b_low − θ_low
        end if
    end if
end for
```

**Algorithm 2** Step Forward Encoding (SFE) Algorithm

**Input:** Continuous signal: $c(t)$, SFE threshold: $\theta_{SFE}$,
**Output:** OUT Channel: $s_{out}(t)$

```
Initialize s_OUT(t) ← zeros of length N = length of s(t)
s(t) ← c(0)
b(t) ← s(0)
for each timestep i do
    s(t) ← c(t)
    if s(t) > b(t) + θ_SFE then
        s_OUT(t) ← 1
        b(t) ← b(t) + θ_SFE
    else if s(t) < b(t) − θ_SFE then
        s_OUT(t) ← −1
        b(t) ← b(t) − θ_SFE
    end if
end for
```

**TABLE I** NCS Modified SFE Configuration

| SFE Parameter | Symbol | Value |
|---|---|---|
| Comparator Threshold | $\theta_{comp}$ | 120 µV |
| HIGH Threshold | $\theta_{high}$ | 4 µV |
| LOW Threshold | $\theta_{low}$ | 10 µV |

**Figure 4. Pseudocode of the modified spike encoding scheme. (Algorithm 1)** Modified version of the SFE algorithm implemented in software and hardware. The modified algorithm computes a signal $c(t)$ to generate two output channels $s_{high}(t)$ and $s_{low}(t)$ based on a comparator threshold $\theta_{comp}$, where each channel has its own fixed threshold $\theta_{high}$ and $\theta_{low}$. **(Algorithm 2).** Classical SFE operation using a single threshold $\theta_{SFE}$.

Figure 5A shows the encoding algorithm system block, generating two spike trains on the two channels $\delta_{\pm}$(t) corresponding to HIGH and LOW, respectively. The encoder obtains the sampled 12-bit resolution data $s$(t) from the FPGA ADC at $f_{in}$ = 2 kHz, generating the spike events every timestep. Figure 5B illustrates a simplified representation of the modified SFE algorithm operation. Where a continuous signal highlighted in blue is processed and the SFE generates spike events based on its current amplitude, evaluated by the comparator $\theta_{comp}$, and its variations. The figure also highlights in red the reconstructed signal based on the firing activity from the two SFE channels, showing the discretization effect due to the dynamics of the SFE algorithm.

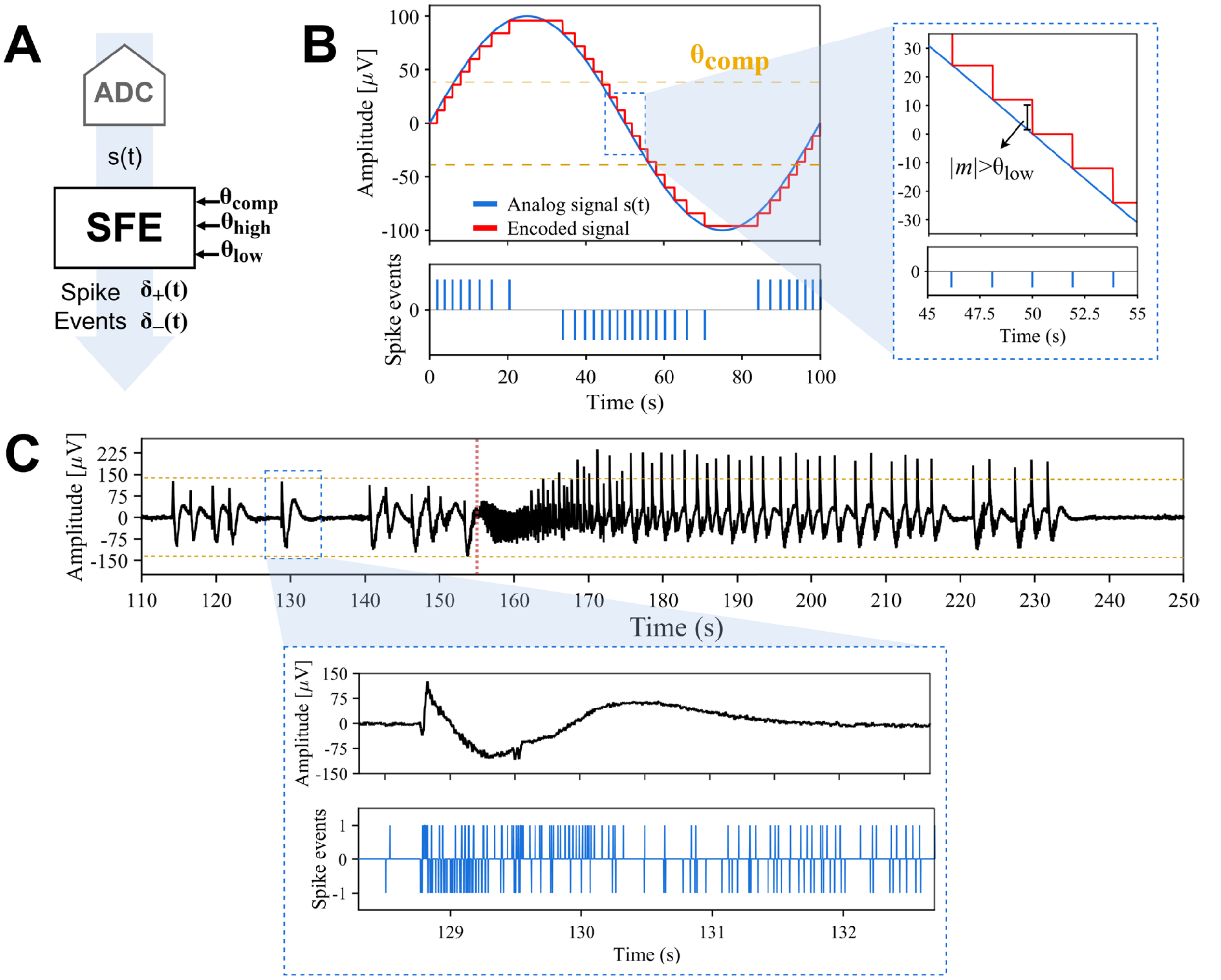

**Figure 5. Modified spike encoding scheme operation. (A)** Modified step-forward encoding (SFE), receiving 3 parameters and generating spike events in two separate channels. **(B)** The algorithm encodes a sampled signal into spike trains, incorporating three different thresholds: $\theta_{comp}$ to divide the signal based on its amplitude, and both $\theta_{high}$ and $\theta_{low}$. Highlighted in yellow is the comparator threshold, which determines in which channel the spikes should be generated. In blue, the analog signal processed by the modified SFE; in red is the reconstructed signal from the spikes generated by the modified SFE **(C)** Example of SFE functionality applied to interictal discharges leading to seizure (onset marked by the red bar). The insert shows the SFE output in response to the interictal event framed by the blue square and consisting of a series of positive/negative spikes, the polarity of which depends on the slope sign of the interictal signal at each computation step-window.

Figure 5C illustrates a representative epileptiform pattern recorded from a hippocampal spheroid, including interictal events and an ictal discharge. The former are recurring short (≤ 2 s-long) electrographic transients as the one framed by the dashed blue box; ictal activity (onset marked by the vertical red dashed line) is a robust (>10 s-long) discharge, of slower recurring rate than interictal events, and consisting of a high-frequency oscillatory pattern (tonic phase) followed by slow-down of network activity (clonic phase) resulting in population bursts recurring at progressively longer intervals until seizure termination. In this case, the ictal discharge is heralded by a barrage of interictal events, which may thus be defined as pre-ictal activity. Interictal and ictal events are separated by epochs of baseline activity, i.e., the signal noise floor devoid of electrographic epileptiform events. To mitigate the risk of high-amplitude interictal discharges generating a disproportionately high spike rate relative to baseline activity, the NCS spike streams are separated according to the comparison threshold to apply a higher SFE threshold for high-amplitude signals. This approach prevents bias in the training algorithm and ensures that the critical influence of baseline activity is preserved, maintaining high accuracy and robustness.

## Hardware FPGA SNN platform operation and description

The hardware implementation of the SNN platform comprises a total of 16 input channels, a reservoir layer of up to 256 spiking neurons randomly interconnected to form recurrent connections within the liquid space, and a readout layer consisting of 16 spiking neurons. These values can be configured by the user before deployment to select the number of active neurons per layer, allowing flexibility in the overall configuration of the SNN.

The neuron model used in the NCS is leaky integrate-and-fire (LIF) due to its simplicity and hardware-friendly implementation. This model follows the dynamics described in Equation (1.a) describes how the activity of presynaptic spikes, $a_i(t)$, together with their synaptic weights, $w_i$, contributes to the membrane potential $V(t+1)$ of a postsynaptic neuron while accounting for a linear leakage decay factor $k$ over time. Only when the membrane potential exceeds a predetermined threshold $\theta_t$, does the postsynaptic neuron fire, producing a spike $s(t)$, after which the potential is reset to a fixed value $V_{reset}$, as expressed in (1.b) and (1.c).

$$\boldsymbol{V(t+1) = V(t) - kt + \sum_{i} (w_i \cdot a_i(t))} \quad (1.a)$$

$$\boldsymbol{s(t+1) = \begin{cases} 1 & \text{if } V(t+1) > \theta_t \\ 0 & \text{otherwise} \end{cases}} \quad (1.b) \quad (1)$$

$$\boldsymbol{V(t+1) = \begin{cases} V_{reset} & \text{if } s(t+1) = 1 \\ V(t+1) & \text{otherwise} \end{cases}} \quad (1.c)$$

The SNN layers use the LIF model described in equation (1). These neurons use quantized integers of 12-bit and 16-bit to model the membrane potential, which is stored in a 0.375KB and a 0,125KB static random-access memories (SRAMs) for reservoir and readout layer, respectively. Both the rest, reset, and leakage parameters are configurable allowing to model multiple dynamics of the neuron behavior in each layer. Similarly, synaptic connections and weights are represented using 8-bit and 12-bit precision and stored in dedicated single port 64 KB and a dual-port 6 KB SRAM (for weight updates purposes during training phase) for both reservoir and readout layers respectively, supporting full customization of connectivity. The reservoir layer has the capacity to allocate 256×256 8-bit connections while the readout has the capacity for 258×16 12-bit weights. The selection of bit width for the system is based on similar real-time platforms and empirical results obtained during the quantization of the hardware implementation in the Python simulator.

Additionally, the NCS implements an online training algorithm, R-STDP, enabling real-time updates of synaptic weights to better adapt its synaptic connectivity based on the ongoing spheroid activity. This could potentially allow the SNN to learn dynamically over different samples of spheroids and their intrinsic unique characteristics. The R-STDP model operates based on principles of the STDP learning process. STDP updates the corresponding synaptic weight change depending on the relative timing between presynaptic spikes (input neuron firings) and postsynaptic spikes (output neuron firings). The changes in synaptic weights are often modelled as an exponential function of the time difference between pairs of spikes. While STDP is traditionally considered as an unsupervised algorithm, R-STDP incorporates the influence of a neuromodulator to regulate the weight updates. These updates are ruled by a label signal, $l(t)$, to combine both STDP and anti-STDP long term potentiation/depression (LTP/LTD) processes during the training phase. Equation (2) shows the dynamics of the R-STDP learning algorithm, where $A_{\pm}$ refers to the amplitude of the weight update, $\Delta t$ refers to the timing between postsynaptic and presynaptic spikes, and $\tau_{\pm}$ is the time constant of the learning curve. Based on the label of the information streamed into the system, the R-STDP algorithm induces STDP (5.a) process in the neuron that should be sensitive to the type of data being streamed into the SNN while the other neuron is punished by an anti-STDP (6.b) process.

$$\Delta w_{\mathrm{STDP}} \begin{cases} \Delta w^{+} = A_{+} \cdot e^{\left(-\frac{|\Delta t|}{\tau_{+}}\right)} & \text{if } \Delta t > 0 \\ \Delta w^{-} = A_{-} \cdot e^{\left(-\frac{|\Delta t|}{\tau_{-}}\right)} & \text{if } \Delta t < 0 \end{cases} \qquad (2.a)$$

$$(2)$$

$$\Delta w_{\mathrm{antiSTDP}} \begin{cases} \Delta w^{+} = -A_{+} \cdot e^{\left(-\frac{|\Delta t|}{\tau_{+}}\right)} & \text{if } \Delta t > 0 \\ \Delta w^{-} = -A_{-} \cdot e^{\left(-\frac{|\Delta t|}{\tau_{-}}\right)} & \text{if } \Delta t < 0 \end{cases} \qquad (2.b)$$

Figure 6 shows the hardware architecture of the learning algorithm accelerator. The NCS implements a series of 6-bit counters per neuron within the readout layer (for both postsynaptic and presynaptic neurons), resetting whenever a new spike is fired to compute the time from the last firing spike. The value from these counters is used to compute both $\Delta W_{\mathrm{STDP}}$ and $\Delta W_{\mathrm{antiSTDP}}$ in a series of look-up tables (LUTs), approach used in similar STDP-based algorithms (22, 23) thereby reducing the computational complexity associated to the weight update. Furthermore, this LUT-based approach enables flexibility in varying learning parameters and curve shapes, supporting a range of applications. The LUTs are implemented to store 64 values of 8-bit signed fixed precision (64 values for each LTP/LTD) processes in a series of SRAMs of 0.5 kb. Given the input sampling interval of 0.5ms ($f_{in}$ = 2 kHz), the LUT table can capture and relate temporal activity over a timespan of up to 32ms. Based on the learning mode, it is possible to configure the NCS to operate with both STDP or R-STDP algorithms, reading from the specific STDP/antiSTDP LUT considering the current neuron being updated. An 8-bit adder adds both positive and negatives updates into Δw, which is then added to the current value of synaptic weight updating it back to the synaptic weight SRAM.

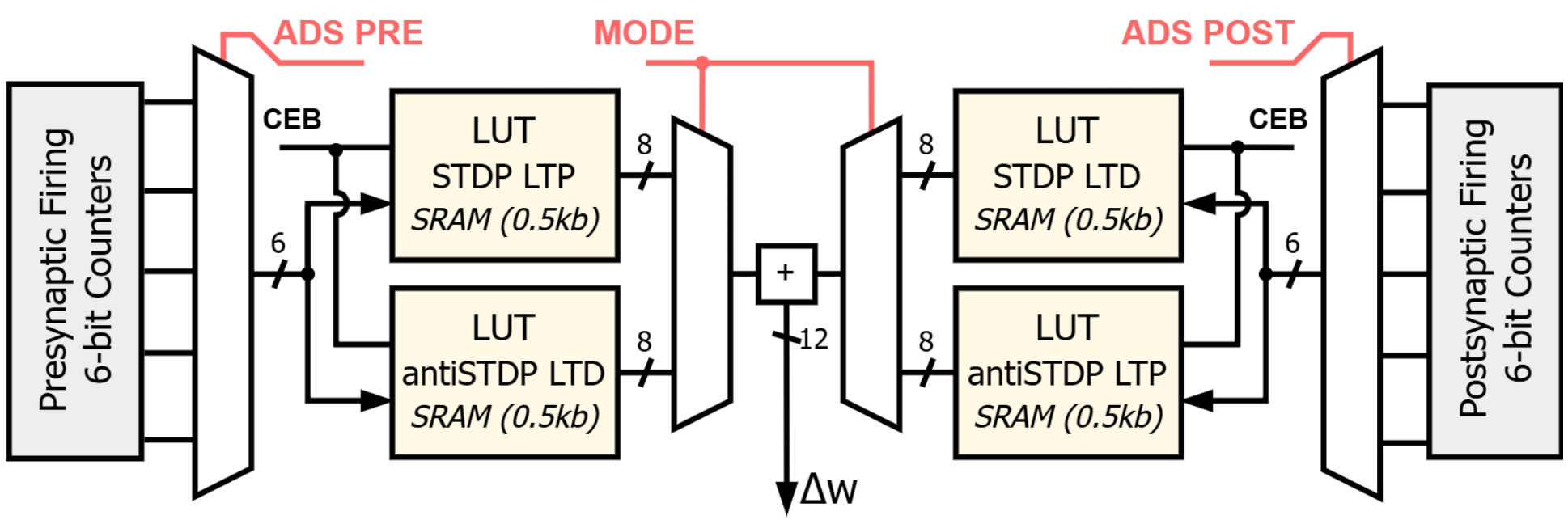

**Figure 6. Learning algorithm hardware architecture.** The system is composed of 4 LUT to store all combinations for LTP/LTD curve functions from both STDP and antiSTDP processes. 6-bit counters associated to the firing of each neuron in the readout layer is used as the address of these LUTs, computing the time between spikes. At the end, both LTP and LTD processes are added, computing the total synaptic update for a single synapse.

The NCS monitors and processes the spheroid's electrical activity in real-time to trigger the activation of the stimulator according to the resulting forecasts. The NCS processes all incoming spiking activity from the input streams and propagates it through the liquid state at each input sampling interval ($f_{in}$ = 2 kHz), ensuring that the system processes the information and generates the corresponding output activity before the next batch of data arrives. The system operates synchronously to a system clock of frequency $f_{sysclk}$ = 100 MHz. At the start of each input sampling interval, the NCS processes the input activity and starts its operation by updating the membrane potential of all neurons in each layer by sequentially iterating over the input channels and all the respective pre-synaptic neurons associated with the neuron being updated. The updates are governed by the accumulation of synaptic weights, from the respective SRAM, associated with the firing activity of pre-synaptic neurons at the time of the processing. When the learning is enabled, the label is transmitted along the input data, indicating which neuron is sensitive to the data sample to choose between the LUT pairs once the neuron is iterated. Before a synaptic weight is read, the system checks the current presynaptic and postsynaptic firing activity, determining if they require any LTP or/and LTD process. At the same time as the synapse is read from the SRAM, the system also reads the corresponding synaptic updates from the learning LUTs, calculating the total synaptic update. In the next cycle, and at the same time the next synapse is about to be read, the system updates the previous one with the new value, since the synaptic weight SRAM is implemented using dual-port SRAM. This implementation reduces the need to wait to write a synapse before reading the next one, speeding up the processing of the network.

This architecture employs time-division multiplexing, allowing a single neuromorphic processing unit (NPU) to sequentially handle the processing of all neurons within the layer. The NPU models the dynamics of the neuron model (activation function) of the neural network. By sharing the NPU circuit across the network we reduce the total number of resources, and the leakage power derived from them.

At the end of its operation, the NCS generates a sequence of spikes from both the reservoir (to be processed in the next interval by the readout layer), and for the readout neuron (which are then used to determine the outcome of the classification process). Since both layers operate simultaneously, the overall time to process a batch of data is dictated by the slower layer. This latency depends on the configuration selected within the platform (the number of active neurons and the sparsity of the reservoir connections) and the activity of the input system. Although the reservoir layer employs sparse recurrent connectivity, its larger scale typically results in higher computational demand compared to the readout layer, making it the dominant factor in determining the system's per-sample processing time. The processing time $p_t$ of the SNN from the NCS can be computed as Equation (3), where $N_n$ refers to the number of neurons in the reservoir layer and $t_{clk}$ the period of the clock signal. Since the NCS, for this application, process input data in samples of 2 kHz ($T_{in}$ = 500 µs) its required to configure the network to a value of $N_n$ such that $p_t < T_{in}$.

$$p_t = (N_n^2 + 4N_n)\, t_{clk} \qquad (3)$$

The hardware model of the NCS is developed using Python 3.11.9 in conjunction with NEST Simulator v3.6.0 and its modelling language NESTML. This approach ensures that the models of the neuron, the learning algorithm, and the dynamics of the NCS meet the constraints imposed by digital circuitry. Finally, hardware simulations were conducted using Xilinx Vivado 2022.2 and implemented on a Zybo Z7 SoC FPGA development board.

## NCS custom deployment configuration on FPGA

The NCS implemented in FPGA is a configurable platform designed to support multiple configurations tailored to different dynamics. To determine the configuration parameters of both the reservoir and readout layers, we performed different analysis to validate the firing dynamics of the NCS system. First, to select both the SFE and reservoir parameters, we computed the inter-spike interval (ISI) analysis of the input spikes at the input to generate consistent spikes for each segment of the dataset used for training. The final configuration used for the reservoir and readout layer are disclosed both in TABLE II, TABLE III, and TABLE IV.The NCS has been configured to just use two input channels connected to 128 active reservoir neurons through sparse synaptic connections controlled by both sparsity parameters $c_{inp\text{-}e}$ and $c_{inp\text{-}i}$ (for excitatory and inhibitory reservoir neurons). The 128 reservoir neurons have been recurrently interconnected according to the sparsity parameters $c_{ee}$, $c_{ei}$, $c_{ie}$, $c_{ii}$, with weights initialized from a Gaussian random distribution with mean $\mu$ and standard deviation $\sigma$. After weight randomization, the synaptic weights are scaled based on the type of connectivity by their corresponding weight ratio parameters $w_{inp\text{-}e}$, $w_{enp\text{-}i}$, $w_{ee}$, $w_{ei}$, $w_{ie}$, $w_{ii}$.

**TABLE II.** NCS Reservoir size/ratio

| Parameter | Value |
|---|---|
| Input channels | 2 |
| Reservoir size | 128 |
| Excitatory ratio | 0.8 |

**TABLE III.** NCS Reservoir synaptic connectivity and weight distribution

| Connectivity Sparsity | | | | | | Weight Distribution | | Weight Ratio | | | | | |
|---|---|---|---|---|---|---|---|---|---|---|---|---|---|
| $c_{inp\text{-}e}$ | $c_{inp\text{-}i}$ | $c_{ee}$ | $c_{ei}$ | $c_{ie}$ | $c_{ii}$ | $\sigma$ | $\mu$ | $w_{inp\text{-}e}$ | $w_{inp\text{-}i}$ | $w_{ee}$ | $w_{ei}$ | $w_{ie}$ | $w_{ii}$ |
| 0.05 | 0.05 | 0.05 | 0.1 | 0.2 | 0.1 | 0.05 | 0.5 | 280 | 150 | 120 | 75 | -175 | -25 |

**TABLE IV.** FPGA LIF Neuron parameters

| Parameter | Reservoir Layer | Readout Layer |
|---|---|---|
| No. of neurons | 128 | 2 |
| $V_{rest}$ | 0 mV | 0 mV |
| $V_{reset}$ | -100 mV | - 2048 mV |
| Threshold $\theta_t$ | 300 mV | 8000 mV |
| Leak $k$ | 1 mV | 10 mV |

Similarly, the readout layer is implemented as a feedforward layer connecting the selected neurons from the reservoir (64 random reservoir neurons) to the two output classification neurons. The synaptic weights in this layer are also randomly initialized using the same random distribution as the reservoir layer; however, the scaling factor used is larger. This increases both synaptic weight magnitude and membrane potential relative to the reservoir. By increasing these magnitudes, the step size of weight updates is reduced, which prevents the weights from exploding and acts as a built-in learning rate mechanism, similar to those employed in other learning algorithms. The specific parameter values used in the readout layer for both inference and learning phases are summarized in TABLE V.

**TABLE V.** Readout Layer connectivity and learning parameters

| Mode | Parameter | Value |
|---|---|---|
| Inference | $\sigma$ | 0.05 |
| | $\mu$ | 0.5 |
| | *Excitatory ratio* | 0.7 |
| | *Weight ratio* | 400 |
| Training | $A_+$, $A_-$ | 10 |
| | $\tau_{\mathrm{pre}}$, $\tau_{\mathrm{post}}$ | 12 |

## Closed-loop electrical stimulation driven by the NCS

Square biphasic current pulses (100 µs/phase, positive phase first) were delivered in monopolar configuration through a selected MEA electrode. We chose the current intensity based on a 'fast' input/output (I/O) curve starting at 150 µA, aimed at finding the stimulus intensity that would reliably (≥80% probability) evoke an interictal-like response (25) without triggering after discharges (i.e., population bursts > 2 s (26)).

The bidirectional communication between the NCS and the spheroid coupled to the MEA system relied on a custom interface PCB reading the MEA signals from the MCS SCU analog output via the SCU 68-pin analog output connector and feeding them to the NCS; a custom graphical user interface enabled selecting the feedback electrode. The NCS forecasts were then sent as TTL pulses (0-3.3 V) back to the custom interface PCB through which they were sent to the selected digital input of the MCS IFB v3.0 via its dedicated digital I/O 68-pin connector. The TTL pulses were then detected by the Digital Event Detector via the Multichannel Experimenter software, in turn activating the built-in stimulator of the MEA system; they were detected by the MEA system at their rising edge, thus requiring one sampling interval (200 µs) to activate the built-in stimulator. To prevent activation of the stimulator by detection of stimulation artifacts, we used the built-in blanking feature of the MEA system. The TTL pulses were also sent to the analog input of the MEA system for real-time visualization and recording on the PC hard drive.

For each spheroid, we selected the feedback electrode from those showing electrical activity based on the signal-to-noise ratio, and we chose the stimulation electrode from those surrounding the feedback electrode based on the reliability and propagation of evoked responses during the fast I/O step.

Pre- and post-stimulus recordings were pursued for at least 20 minutes to ensure that the final recording segment (from the onset of the first to the end of the last recorded ictal event) was aligned with the 20 min duration of the NCS-driven stimulation session for a robust comparison of the %time in ictal state across the experimental phases. We chose a stimulation session duration of 20 minutes based on the average interval between ictal discharges, so to be ≥3 times the interval observed during pre-stimulus baseline, coherent with the guidelines of the International League Against Epilepsy on the definition of seizure freedom (27).

The NCS was tuned manually during the pre-stimulus baseline recording, starting from a list of available parameter values established empirically during the testing phase of the NCS against pre-recorded MEA signals. The NCS behavior had to meet the following specifications: (i) it would respond primarily during the signal noise floor (forecasting) rather than to the epileptiform events (detection); (ii) it would never respond solely to ictal activity (seizure detection); (iii) if an ictal discharge or an afterdischarge would occur, it would not stimulate during their initial (tonic) phase since the biological network is in a non-permissive state and it would not be possible to entrain it via electrical stimulation.

## Data and statistical analysis

Epileptiform discharges were labeled by expert neurophysiologists using a semi-automated approach based on an automated event detection algorithm followed by inspection and manual correction of the obtained labels, as required. The software for event labeling, running in a graphical user interface, was written in MATLAB R2021b (MathWorks, Natick, USA).

To evaluate the degree of ictal activity reduction, we computed the % time spent by the spheroids in the ictal state, as described in (28). We adopted this approach instead of measuring the duration and inter-event interval of the ictal discharges to account for the possibility of only observing one or two ictal events during electrical stimulation, which would bias the quantification of the degree of ictal activity reduction.

For statistical comparison of the %time spent in the ictal state, we first checked the dataset for normality (Shapiro-Wilk test) and homoscedasticity (Levene test). Since the dataset was neither normally distributed ($p = 0.002$) nor homoscedastic ($p = 0.02$), we used one-way ANOVA followed by the non-parametric Games-Howell post-hoc test for multiple comparisons. We considered differences to be statistically significant if $p < 0.05$. To compute the statistical power achieved, we performed a posteriori analysis using the freely available software G*Power.

Throughout the text, data are expressed as mean ± SD, unless otherwise specified.

# Results

## Step forward encoding and reservoir liquid discriminative capability

The modified SFE serves two main purposes: first, to generate a continuous stream of spike trains in real-time at an input sampling frequency of $f_{in} = 2$ kHz which can be processed by the NCS; second, to capture the dynamics of signal variations and activity, enabling efficient temporal feature extraction by the reservoir. To configure the modified SFE algorithm and compare its response to the classical one (both already described in the Materials and Method section) We conducted an ISI analysis and assessed the total activity generated by both SFE variants, comparing their outputs across ictal and non-ictal labeled recording segments.

Figure 7 presents a comparison of spike distributions between labeled “ictal” and “non-ictal” signal recordings from the MEA systems. Figure 7A and 7B display representative epileptiform patterns recorded from a hippocampal spheroid, illustrating a purely interictal epileptiform phenotype and a mixed interictal-ictal phenotype, respectively. Figure 7C and 7D show the response of the classical SFE algorithm. Using this method, ictal events yield the generation of a higher number of spikes compared to non-ictal segments, which may produce bias during the training process. Moreover, the spike density distribution is less pronounced, with both HIGH and LOW channels of ictal-labeled activity displaying similar activity. In contrast, during interictal activity, spikes are more evenly and differentially distributed across multiple timesteps. Figure 7E and 7F present the evaluation of the modified SFE algorithm.

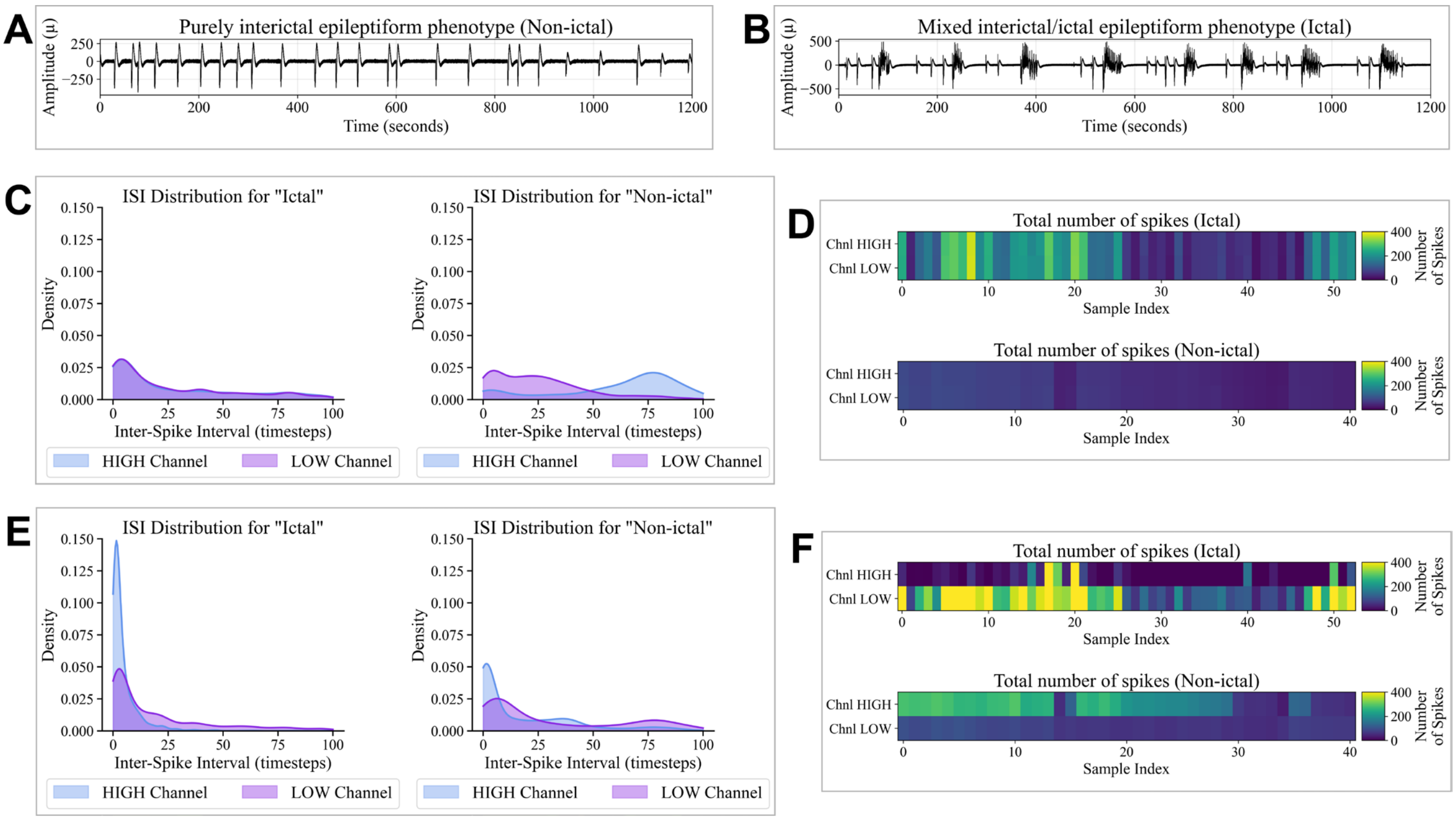

**Figure 7. Firing rate and ISI analysis of the SFE encoding algorithm and the proposed modified SFE. (A-B)** Representative epileptiform patterns recorded from the hippocampal spheroids, where **(A)** corresponds to those patterns related to non-ictal events and **(B)** to ictal events. Panels **(C-D)** present the firing analysis of the input channels the traditional SFE, which provides limited distinction between ictal and non-ictal samples. **(E-F)** illustrates the analysis of the same input channels using the modified SFE. Here, each channel exhibits a distinct firing response, suggesting that the modified SFE can capture diverse patterns and features. These extracted characteristics can subsequently be leveraged by the reservoir and classification layers to improve seizure forecasting.

As can be observed from the ISI analysis, most spikes are concentrated within a short time window (0 to 20 timesteps) during ictal activity. This temporal clustering is expected to have a stronger influence on the SNN due to the leaky dynamics of the neuron model, generating more spikes within the network. Additionally, the proposed modified SFE produces a clearer differentiation in spike density distributions, especially for ictal-related events. Finally, the total spike count over the segments is less pronounced. This suggests a more balanced and distributed encoding, which may lead to improved performance during the training process. The reservoir layer of the NCS is responsible for extracting the temporal features embedded in the streamed input data, enabling subsequent classification by the R-STDP learning algorithm implemented in the readout layer.

Figure 8 illustrates the spiking activity analysis of the NCS during real-time processing of streamed data. Figure 8A and 8B show the spiking activity of the reservoir following the presentation of the SFE streamed spikes in two examples of segments classified as ictal and non-ictal respectively. SFE HIGH and LOW channels, highlighted both in blue and violet respectively, generate a spiking response in the reservoir dynamics. All reservoir neuron spikes are shown in light grey, while the sublet of neurons designated as outputs of the liquid state (the ones connected to the readout layer) are highlighted in black. These output neurons are not directly connected to the input channels; instead, their activity arises from the internal recurrent dynamics of the reservoir, ensuring the extraction of temporally rich, high-dimensional representation of the input dynamics. Figure 8C displays the total number of spikes generated by the samples of input data during the training process, while Figure 8D shows the deviation of firing activity for both types of segments.

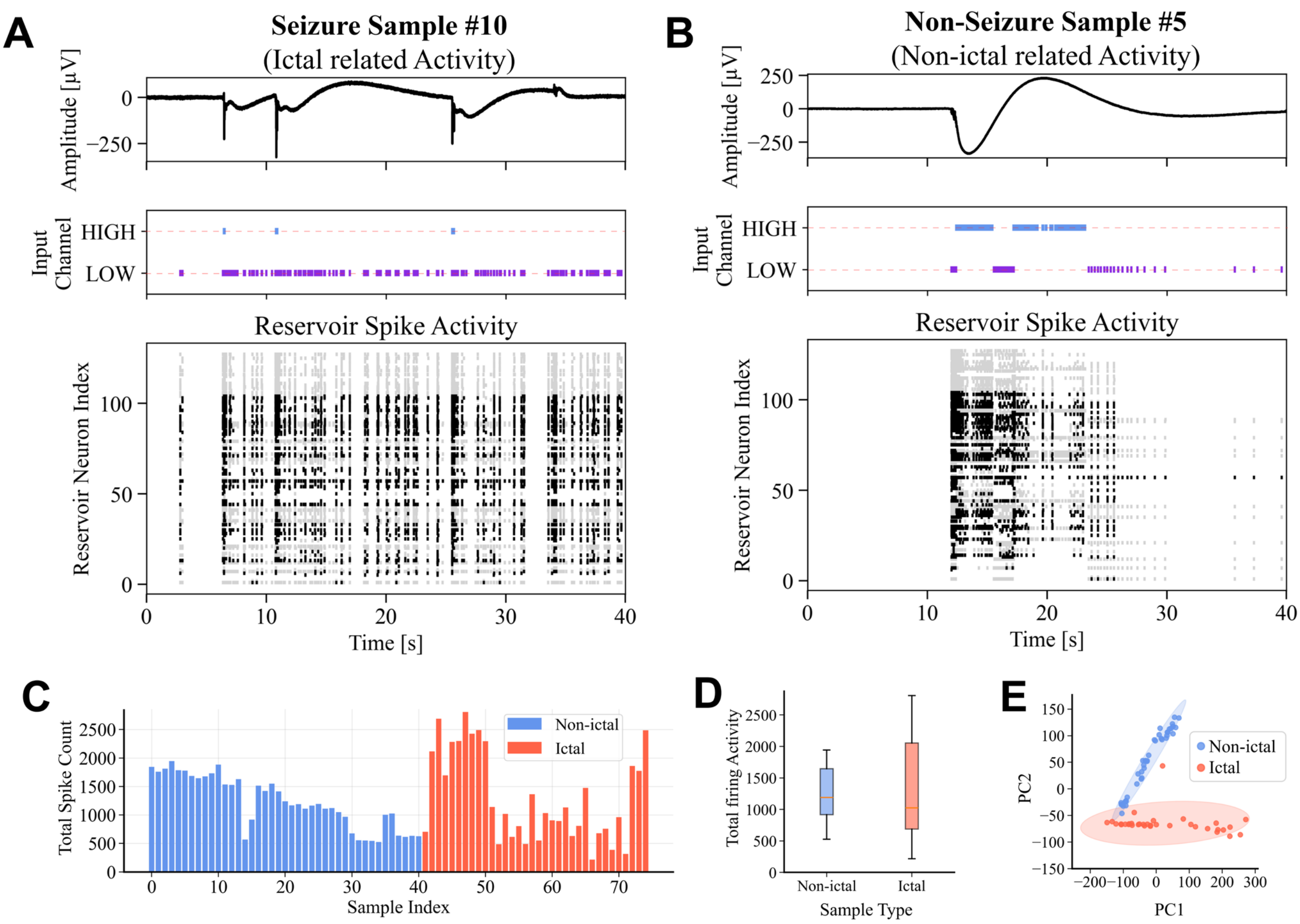

**Figure 8. spike analysis of both input stream data and reservoir activity. (A-B)** Liquid network spiking activity over different samples of input stream data, highlighted in black the activity of those neurons that compose the liquid state connected to the readout layer. **(C)** Spike rate distribution over the segments used for the training process. **(D)** Distribution of the spike rate over both types of labeled segments. **(D)** PCA analysis over the reservoir liquid state, validating the NCS ability to cluster data so it can be linearly classified by the readout layer.

The number of reservoir spikes ranges from approximately 2500 down to a minimum of 400 spikes, reflecting a spike sparsity of 93.75% to 99%, which reduces the amount of data the NCS is required to process during the inference process. We also conducted a principal component analysis (PCA) on the spike count vectors corresponding to each output liquid neurons to assess the quality of the reservoir´s initialization in conjunction to the SFE encoding algorithm. Figure 8E illustrates the resulting PCA projection of the two input classes, showing two well-separated clusters corresponding to each class. This clustering demonstrates that the preprocessing layer composed of the SFE algorithm, and the reservoir dynamics were properly initialized. This could allow the system to generate separable representations during the preprocessing stage of the NCS operation and induce stimulation to prevent a forecasted ictal event.

## Offline training using pre-recorded activity from MEA-coupled hippocampal spheroids

Figure 9 shows the performance of the SNN during the training process. Figure 9A shows the distribution of synaptic weights connecting the reservoir network to the readout layer. The plot highlighted in blue illustrates the distribution of synaptic weights connected to the neuron sensitive to "ictal" labeled segments, while the plot in red indicates the distribution of weights for the "non-ictal" sensitive neuron. Figure 9B uses the same color scheme, showing the distribution of weights after the training process consisting of 50 epochs. The updated weights exhibit a bell-shaped distribution, indicating that most of these connections have little to no effect over the activation function of the readout neurons. Such behavior suggests that the readout layer has learnt to recognize the presynaptic neurons capturing relevant dynamics to classify the type of sample processed by the reservoir. Figure 9C illustrates the evolution of synaptic weights from the readout layer over the epochs, showing the variation of the synaptic weight distributions during the training process. The accuracy of the system during the training process is shown in Figure 9D, where the NCS starts with an accuracy of approximately 10%, reaching up to 96% accuracy at the end of the training process.

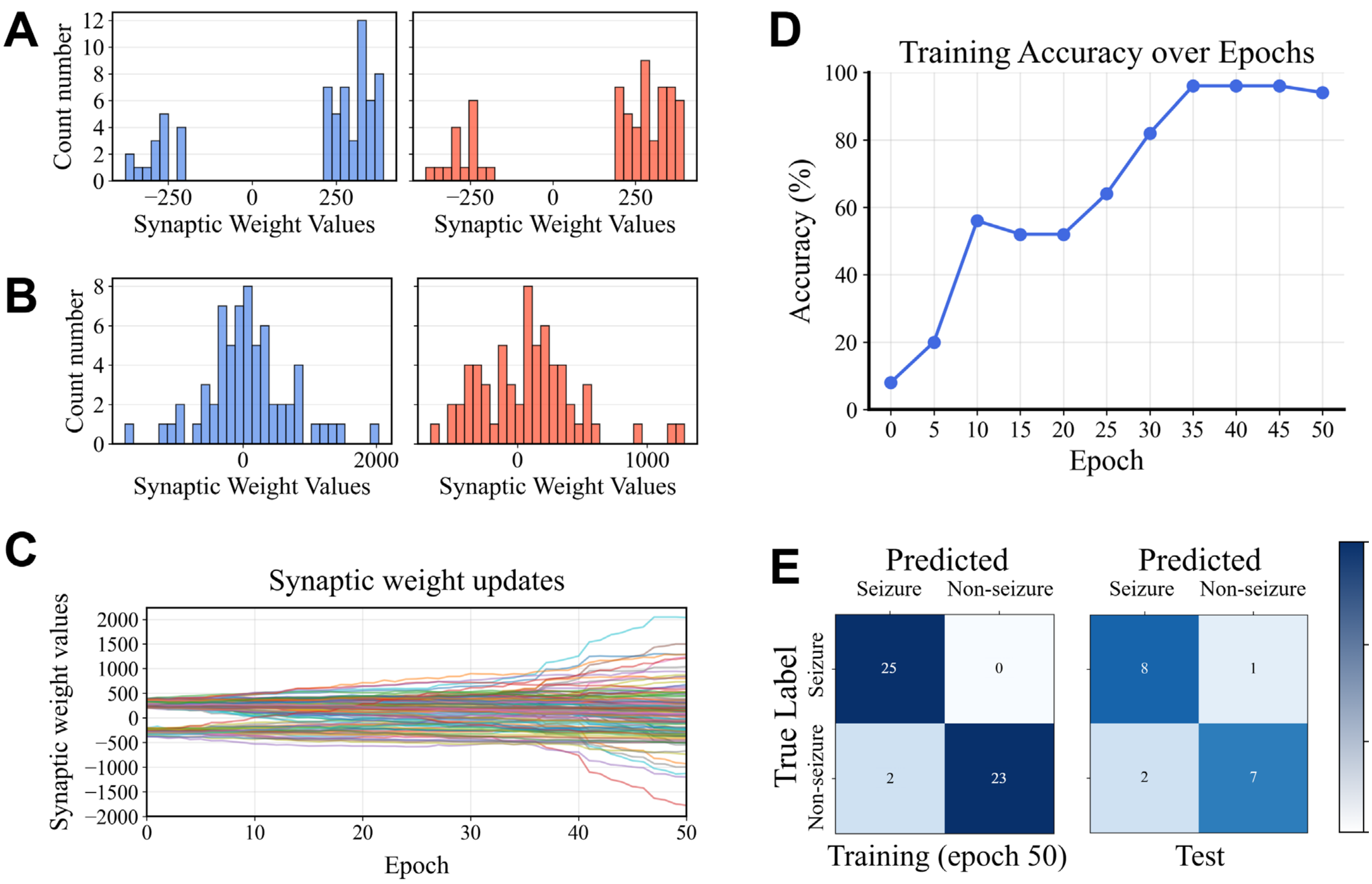

**Figure 9. Training and test process results during the offline training. (A)** Initialization of the synaptic weights using a gaussian distribution (in blue the neuron sensitive to ictal related activity and in red the one sensitive to non-ictal related activity). **(B)** Synaptic weights distribution at the end of the training process, the plots follow the same color scheme as (A). **(B)** Evolution and dynamic update of synaptic weights from LTP/LTD R-STDP updates during the training phase. **(D)** The performance of accuracy of the NCS during the training phase. We evaluated the system every 5 epochs using the training set. **(E)** Confusion matrices of the evaluation of the NCS after training process with both the training set (on the left) and the test dataset (on the right).

In the context of neurostimulation and seizure forecasting, it is important to address the performance of the system with both the accuracy and the ratio of false positive/negatives (FP/FN). Figure 9E shows both these results in the form of a confusion matrix. The matrix on the left shows the ratio at the end of the training phase over the training set, while the one on the right corresponds to the test phase; for the latter, we used a test set comprising 16 20-second signal segments that were not used during training, thereby avoiding bias in the performance evaluation.

Our design achieves a test accuracy of 83.33%, with only one ictal segment misclassified as non-ictal. From both confusion matrices (training and test phases), we observe how the NCS is more sensitive in classifying segments as ictal than non-ictal, resulting in a higher number of FP compared to FN. During training, the NCS successfully classified most samples except for two non-ictal segments, which were classified as seizure occurrences (FP). During the test phase, the NCS failed to distinguish 3 signal samples, two as FP and one as a FN. We consider that such behavior is acceptable, prioritizing stimulation in case of uncertainty, thereby reducing the risk of missing an actual ictal event.

## On-line validation of NCS-driven stimulation in MEA-coupled hippocampal spheroids

To validate the NCS's ability to perform closed-loop control of seizure activity, we employed the same hardware architecture depicted in Figure 1, this time activating the control over the MEA system's stimulator via TTL signals generated by the NCS upon positive seizure forecasts with real-time stream data from hippocampal spheroids.

Figure 10 illustrates the experimental setup used in the laboratory, composed of the NCS system and the PCs used for collecting the information and verifying the operation of the system in real time. TABLE VI summarizes the NCS hardware specifications. The NCS has a total of 272 neurons, 70K synaptic connections, and a total memory of 75KB approximately. Using the full capacity of the NCS running at 100 MHz, the processing speed would result in 665,6 µs/sample, insufficient for the processing speed required of this application running at 500 µs/sample. Therefore, for these experiments, the FPGA was configured to a 2×128×2 topology, leading to a processing speed of 168 µs/sample. The results of power consumption and resource utilization are also detailed in TABLE VI, with values obtained from the summary of the implementation in Vivado.

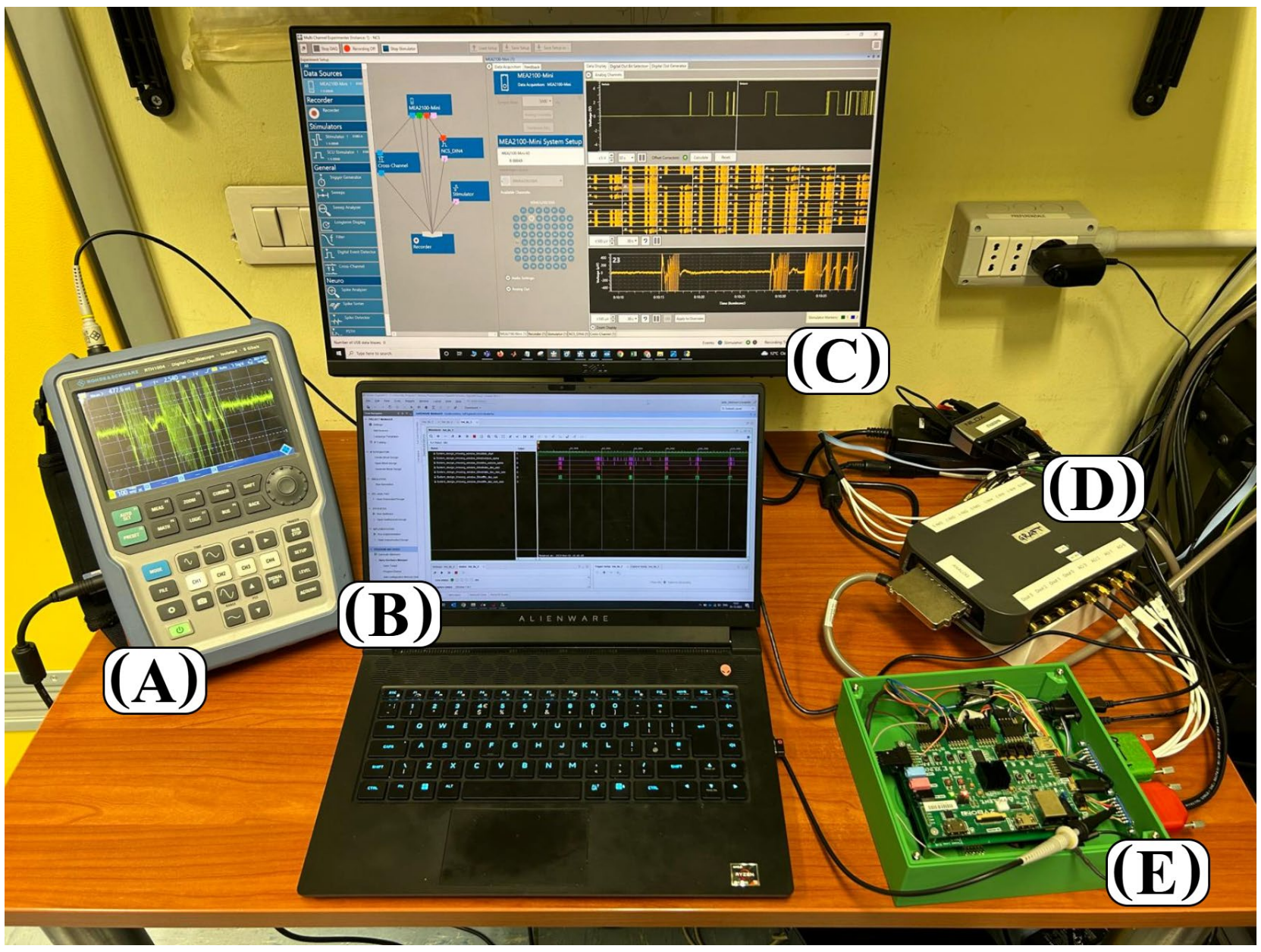

**Figure 10. NCS experimental setup for processing hippocampal spheroids in real-time (A)** Oscilloscope to validate the range of the MEA signals fed into the FPGA. **(B)** Vivado and the integrated logic analyzer (ILA) to corroborate the operation of the FPGA in real time. **(C)** MEA software system to capture the results for the FPGA and the spheroid response over the stimulation **(D)** MEA driver and interface PCB to feed the custom AFE circuitry. **(E)** FPGA Zybo Z7 development board implementing the NCS.

**TABLE VI** NCS hardware specifications

| FPGA Board | Zybo Z7: Zynq-7000 | | |
|---|---|---|---|
| Total # of neurons | 16 × 256 × 16 | | |
| Total # of synapses | 65526 (res)<br>4096 (readout) | | |
| Total Memory | 64,375 KB (res)<br>6,375 KB (readout) | | |
| Learning | STDP, R-STDP | | |
| Clock Frequency | 100 MHz | | |
| Processing speed | 665,6 µs/sample | | |
| Power consumption | 1.7W | | |
| Resources Utilization | LUT | 21347 | (40%) |
| | LUTRAM | 3347 | (19.42%) |
| | FF | 28128 | (26.44%) |
| | BRAM | 121 | (86.43%) |
| | DSP | 2 | (0.91%) |

Figure 11A illustrates the experimental design for this set of experiments (see Materials and Methods for full details). Figure 11B shows a hippocampal spheroid coupled to a 3D MEA and the selected feedback and stimulating electrodes for the representative experiment illustrated in Figure 11C. The latter demonstrates the ability of the NCS to effectively drive electrical stimulation to prevent ictal activity. Upon stimulus withdrawal, seizure activity emerged back, confirming that its disappearance during stimulation was due to the NCS actions rather than by chance. Furthermore, the overall electrical pattern of the spheroid post-stimulation appeared to have slightly changed compared to pre-stimulus baseline, suggesting a possible neuromodulation effect of the NCS-driven stimulation. The insets on the right in Figure 11C show the signal segments highlighted in the recording overview, at an expanded time scale, to emphasize ictal events recorded pre- and post-stimulation as well as the patterned stimulation driven by the NCS.

Overall, we have collected $n = 15$ validation experiments, using unique spheroids. The biological sample population was obtained from three litters (Litter 1: 3 spheroids; litter 2: 2 spheroids; litter 3: 10 spheroids). Results consistently demonstrated the dramatic reduction of ictal discharges by the NCS-driven stimulation (Figure 11D; %time spent in ictal state – pre-stimulus: 23.23 ± 11.77%; NCS-driven stimulation: 0.59 ± 1.15%; post-stimulus: 25.8 ± 12.22%; one-way ANOVA, F(df): 22.43(2), $p < 0.0001$; Games-Howell post-hoc test: $p < 0.0001$ stimulation vs pre- and post-stimulation, $p = 0.99$ pre- vs post-stimulation; effect size: 0.96; power: 0.99). Analysis of the NCS-driven pulse timings unveiled that the NCS delivered stimuli across a wide range of instantaneous frequencies (0.02-714.3 Hz) consistent with its adaptive behavior, but, remarkably, it primarily operated in the low-frequency regime (Figure 11E). Specifically, instantaneous frequencies within 20 Hz occurred with ~80% probability with a large portion of the pulses (~73% probability) delivered at 10 Hz or less, whereas peak frequencies ≥100 Hz were sporadic (~3.4% probability).

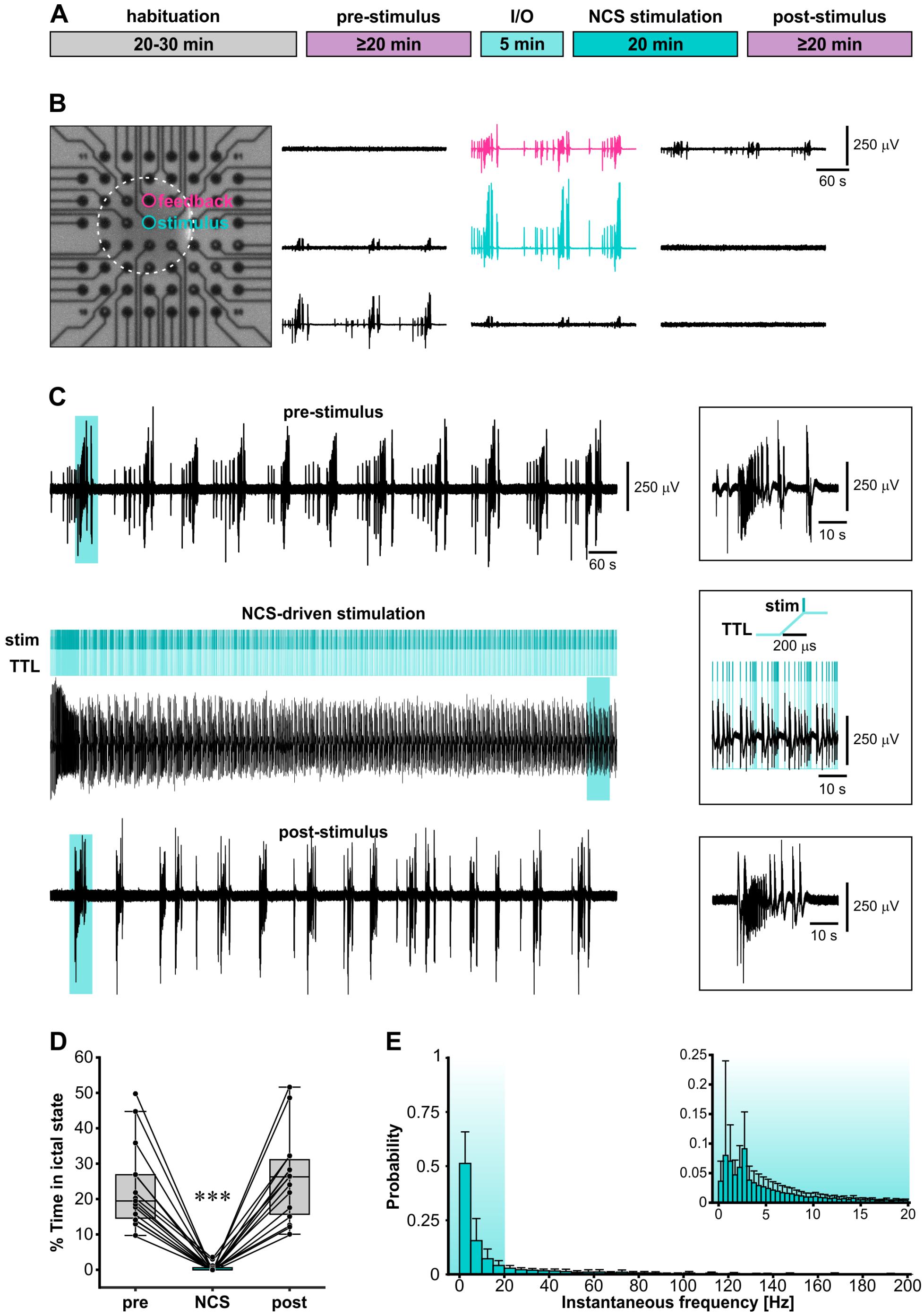

**Figure 11. Seizure activity reduction by NCS-driven patterned stimulation. (A)** Phases of the experimental protocol. **(B)** Optical image of a hippocampal spheroid (outlined by the dashed white circle) coupled to a 3D MEA and 5-minute representative signals recorded by the electrodes coupled to the spheroid. The selected feedback and stimulation electrodes are color-coded as in the optical image. **(C)** Representative experiment using the spheroid in (A) showing the typical pattern of recurring interictal and ictal events generated by the spheroid during pre-stimulus baseline. NCS-driven stimulation completely prevented ictal discharge generation (pulse timings are indicated by the vertical turquoise bars). Upon stimulus withdrawal, the spheroid started generating ictal discharges again. While the overall % time in ictal state (19.5%) was similar to what measured during pre-stimulus baseline (24.5%), the overall spheroid electrical pattern appeared slightly modified. In the NCS-driven stimulation subplot, vertical bars indicate the NCS-generated TTL pulses (TTL) and the electrical stimuli activated by them (stim). The insets on the right of each experimental phase show the signal segments corresponding to the turquoise shaded areas, visualized at a faster time scale. Note the patterned stimuli delivered by the NCS, evoking short population responses resembling interictal events. Above the stimulation inset is a representative segment of NCS TTL output and the triggered stimulus, recorded through the MEA system. The time lag of 200 μs corresponds to one sampling interval, confirming the real-time operation of the coupled NCS-MEA systems. **(D)** Quantification and results statistics for the entire dataset (n = 15 experiments) demonstrates a statistically significant reduction of the % time in ictal state by the NCS-driven stimulation (*** p = 0.0001). Each dot represents an experiment. Solid lines connect the experimental phases to illustrate the trend of each experiment. **(E)** Probability distributions of the instantaneous frequency of stimulations delivered by the NCS, averaged across the experimental dataset (bin size = 5 Hz): the NCS primarily operates at instantaneous frequencies <= 20 Hz. The inset shows the probability distribution of instantaneous frequencies within 20 Hz for bin size = 0.5 Hz to further emphasize the predominance of low-frequency stimulation. Data are expressed as mean ± SD.

## Comparison with State-of-the-Art

TABLE VII compares our design with related works. Because FPGA-based real-time seizure prediction systems are still scarce, we also include seizure detection implementations to provide context on FPGA resource usage and power consumption for similar real-time applications.

**TABLE VII.** Comparison with the state-of-the-art on FPGA real-time devices

| | | (30) | (31) | (32) | (33) | (34) | **This work** |
|---|---|---|---|---|---|---|---|
| **FPGA** | | Zynq-7000 | Cyclone IV | Zynq-7000 | Virtex-7 | N/A | **Zynq-7000** |
| **Network** | | SVM | ANN | DNN | ANN | SVM | **LSM** |
| **Dataset** | | CHB-MIT | TUH EEG Corpus | AES Seizure Prediction | CHB-MIT | CHB-MIT | **Hippocampal Spheroids** |
| **Accuracy** | | 98.4% | 95% | 74% | 87.4% | 64.9% | **83.33%** |
| **Frequency (MHz)** | | 100 | 50 | 25 | 20 | N/A | **100** |
| **Power (W)** | | 0.380 | N/A | 1.909 | 0. 218 | 0.009 | **0.080** |
| **FPGA Resources** | **LUT** | 11390 | 12971 | 12910 | N/A | N/A | **4650** |
| | **LUTRAM** | N/A | N/A | 2956 | N/A | N/A | **158** |
| | **BRAM** | 62 | N/A | 60 | N/A | N/A | **41** |
| | **FF** | 11748 | 114 | 10621 | N/A | 3210 | **5342** |
| **Application** | | Seizure detection | Seizure detection | Pre-ictal detection | Seizure prediction | Seizure prediction | **Seizure Forecasting** |

Our proposed SNN architecture achieves a seizure forecasting accuracy of 83.33%, while consuming only 80 mW of average power (from Vivado power analysis) and requiring fewer FPGA resources than comparable works. In contrast, Tahar et al. (32) do not report the power consumption of their DNN but the global FPGA power (1.9 W). Our proposed system, summarized in TABLE VI, consumes 1.7 W, which is still 200 mW less power than what reported in (32). Such difference may result from the usage of spiking mechanisms and their reduced computations compared to traditional DL approaches. On the other hand, Taufique et al. (33) achieved just 4% higher accuracy than our design but at the cost of approximately 2.7× higher power consumption. Since FPGA resource utilization is not reported, it is difficult to identify the source of these differences. Finally, the model reported by Alaa et al. (34) consumes a very low average power (9 mW) but in this case the authors implemented a support vector machine (SVM) algorithm instead of a whole network. Nevertheless, the reported accuracy of 64.9% constitutes a significant limitation. This accuracy, and the high sensitivity of 95% reported in their work, could result in frequent false alarms (FP) and potential overstimulation in patients.

Regarding seizure detection, robust results are reported by Wang et al. (30) and Sarić et al. (31), with accuracies of 98.4% and 95%, respectively. Although the application is different and some details of power and FPGA utilization are missing, we can still draw some useful comparisons. For example, in (30) they implement a SVM system to process real-time EEG, consuming 380 mW and using 11390 LUTs and 11748 FFs. Compared with our SNN, we achieve 4.5× less power consumption, and approximately 2.45×-2.2× fewer LUT and FF resources. In (31), the power consumption is not reported, but the system requires 2.8× more LUTs than ours, while only synthesizing 114 FFs. This discrepancy may result from differences in synthesis tool inference (e.g., mapping FFs as LUTs) or from their adoption of a more event-driven architecture that minimizes the need for stored values, leading to globally comparable resource usage.

It is important to note that a direct comparison across all these systems is challenging, since the datasets, preprocessing pipelines, and evaluation protocols differ considerably. As a result, performance metrics such as accuracy cannot be taken as absolute indicators of superiority. Nevertheless, these works provide valuable reference points regarding power consumption and FPGA resource utilization typically reported for seizure prediction and detection systems. Placing our design alongside them offers a sense of where our approach stands in terms of efficiency and scalability, without implying that one solution is universally better than another. Instead, each design highlights different trade-offs that reflect the chosen architecture and experimental setting.

## Discussion

We have demonstrated the potential of a reservoir NCS implementing R-STDP to perform real-time seizure forecasting and consequently drive electrical stimulation to prevent seizure activity. To this purpose, we have leveraged rodent hippocampal spheroids as a high-yield in vitro model of spontaneous hippocampal epileptogenesis (20), which is relevant for addressing the most common DRE, i.e., mesial temporal lobe epilepsy. The spontaneous emergence of epileptiform activity in hippocampal spheroids is a distinctive feature of this model compared to other in vitro or ex vivo preparations, which rely on acute induction of epileptiform activity through pharmacological or ionic manipulation of the recording medium. The standardized culturing protocol described in (20) permits reproducibility of this model, thereby offering a simplified high-throughput testbed for future studies on epileptic syndromes involving the hippocampus. While variability is an inherent aspect of any biological model, the epileptiform behavior of these spheroids offers a diverse range of electrical patterns including ictal (seizure) and interictal (between seizures) discharges that are not artificially induced but reflect the intrinsic epileptogenic propensity of the spheroids. This unique feature of the spheroids brings parallelism with epileptic syndromes of unknown origin. Furthermore, the generated epileptiform events resemble those observed in epileptic animals and in human epilepsy, both in terms of macroscopic signal features and of intrinsic frequency components (20). Thus, we anticipate that the use of the NCS can be extended to in vivo studies in epileptic rodents as a future step toward clinical translation.

The NCS efficacy relies on its ability to robustly capture and decode distinctive features of local field potentials to forecast seizure activity through a computation- and power-efficient algorithm. Remarkably, NCS-driven stimulation achieved an outstanding reduction in seizure activity (>97%, i.e., approximating seizure prevention) while primarily using instantaneous stimulation frequencies within the low-frequency range. These features denote the potential of the NCS as a neuromodulation device supporting longer battery life, reduced electrode deterioration and brain tissue stress compared to current closed-loop devices relying on computationally demanding algorithms and typically operating in high stimulation frequency ranges (35, 36). It is worth noting that while high-frequency stimulation is typically preferred in clinical practice, low-frequency stimulation protocols are still much studied in a continued effort to improve both the therapeutic efficacy and the battery life of the neurostimulation device. Particularly, on-demand low-frequency stimulation has demonstrated a valid approach both in preclinical (37) and clinical (36) most recent studies. The schismatic debate around the benefit of high- versus low-frequency stimulation reflects one major outstanding challenge in neuromodulation, i.e., finding the most effective stimulation parameters (7). In this regard, comparative clinical studies, such as (38), emphasize the importance of tailoring stimulation to the patient’s needs. Still, the current approach to devising personalized neurostimulation therapies relies on on-demand open-loop stimulation using arbitrarily predefined stimulation frequencies, the quest of which pursues a one-size-fits-all solution. In addition, stimulation is delivered upon detection of a seizure (e.g., (39)) or of signal features that, when combined, encode high likelihoods of transition to seizure (40). These approaches only partially meet the definition of personalized adaptive treatment and lend themselves to sub-optimal efficacy. When stimulation is delivered upon seizure detection, the aim of the therapy is terminating rather than preventing the seizure; thus, the patient might still experience clinical symptoms; moreover, the epileptogenic network might not respond to stimulation (11). On the other hand, when stimulation is delivered during high likelihood of transitioning to seizure, it might trigger a seizure beforehand (41, 42). In keeping with these concepts, a recent study has found it more beneficial to stimulate during low likelihood of transition to seizure (15).

The NCS presented here operates in a radically new way by using free-run stimulations that are not constrained by pre-defined stimulus trains. This operating mode inherently supports truly personalized neuromodulation therapies, as the output forecasts of the NCS are directly translated into electrical stimulation. To our best knowledge, this is the first demonstration of such a neuromodulation system. While previous works have explored FPGA-based approaches for seizure prediction and detection (see TABLE VI), they have not deployed real-time closed-loop stimulation. The combination of seizure forecasts and forecast-driven stimulation renders our FPGA-based design unique compared to prior systems. Its demonstration fosters a new mindset in the neuromodulation field by exemplifying the potential of ad-hoc free-run stimulation driven by seizure forecasts to prevent seizure occurrence.

## Limitations and Future Work

Hippocampal spheroids are a reductionist model that does not recapitulate the complexity of whole epileptogenic networks, where multiple brain areas are involved in dynamic transitions to seizure. Nonetheless, their use has allowed us to validate the anti-seizure efficacy of the NCS while respecting ethical concerns about testing such new technologies in behaving animals. This step represents a critical milestone supporting the future validation of the NCS in freely moving epileptic animals.

One limitation of the NCS is the need for manual tuning of its major operating parameters by a knowledgeable operator. While this task is minor compared to the current trial-and-error approach to identifying the optimal stimulation strategy, it remains time-consuming. Future work will focus on equipping the NCS with a machine-learning algorithm for auto-tuning its operating parameters, thereby unlocking its full potential.

### Ethics statement

The procedures involving animals were approved by the animal welfare board of Instituto Italiano di Tecnologia and by the Italian Ministry of Health (refs. 176AA.N.9AU and 176AA.N.UWY).

## Acknowledgments

The authors thank Marina Nanni, Alice De Benedetti and Andrea Arena for their technical support for the spheroid cultures; Dr. Giacomo Pruzzo for technical support with the electrophysiology equipment and for having designed and made the custom heating lid for MEA electrophysiology.

## Funding

This work has been supported financially by the European Union Horizon 2020, FET Proactive (RIA)-2018, HERMES – Hybrid Enhanced Regenerative Medicine Systems, Grant Agreement n. 824164, and by Innovations-Fonden under PIPESENSE project (0224-00056B)

## Author contributions

Maryam Sadeghi[1†], Darío Fernández-Khatiboun[1†], Yasser Rezaeiyan[1†], Saima Rizwan[2], Alessandro Barcellona[3], Andrea Merello[3], Marco Crepaldi[3], Gabriella Panuccio[2*], Farshad Moradi[1*]

- Conceptualization: MS, GP, FM.
- Methodology: MS, DFK, GP, FM
- Software/Hardware: DFK, YR, AB, AM, MC, GP
- Validation: MS, DFK, YR, GP
- Formal analysis: MS, DFK, GP
- Investigation: SR, GP
- Resources: MC, GP, FM
- Data Curation: GP
- Writing - Original Draft: MS, DFK, YR, SR, GP, FM
- Writing - Review & Editing: All
- Visualization: MS, DFK, YR, GP
- Supervision, project administration, funding acquisition: GP, FM

## Competing interests

All authors declare they have no competing interests.

## Data and materials availability

All data, code, and materials will be shared upon reasonable requests from qualified researchers with respect to intellectual property protection.